\documentclass[a4paper, 10pt, onecolumn]{article} 
\usepackage[top=3.5cm, bottom=3cm, left=3cm, right=3cm]{geometry}
\usepackage{hyperref}
\usepackage{url}

\usepackage{graphicx} 
\usepackage{subfigure} 
\usepackage{wrapfig}

\usepackage{natbib}

\usepackage{algorithm}
\usepackage{algorithmic}

\usepackage{hyperref}

\usepackage{amsfonts,amsmath,amssymb}

\DeclareMathOperator{\Acal}{\mathcal{A}}
\DeclareMathOperator{\Dcal}{\mathcal{D}}
\DeclareMathOperator{\Ncal}{\mathcal{N}}
\DeclareMathOperator{\EE}{\mathbb{E}} 
\DeclareMathOperator{\NN}{\mathbb{N}} 
\DeclareMathOperator{\RR}{\mathbb{R}} 
\DeclareMathOperator{\tr}{Tr}
\DeclareMathOperator*{\argmin}{\mathop{\mathrm{argmin}}}

\newtheorem{theorem}{Theorem}[section]
\newtheorem{remark}[theorem]{Remark}

\renewcommand\footnotemark{}
\title{Learning the Structure for Structured Sparsity\thanks{This article corresponds to the version of \citep{SheBac15} accepted for publication in IEEE Transactions on Signal Processing. Copyright (c) 2015 IEEE. Personal use of this material is permitted. Permission from IEEE must be obtained for all other uses, in any current or future media, including reprinting/republishing this material for advertising or promotional purposes, creating new collective works, for resale or redistribution to servers or lists, or reuse of any copyrighted component of this work in other works.}}

\author{
Nino Shervashidze \\
Mines ParisTech, PSL Research University -\\
Centre for Computational Biology\\ 
Institut Curie, INSERM U900\\ 
Paris, France \\
\texttt{nino.shervashidze@mines-paristech.fr} \\
\and
Francis Bach \\
INRIA -\\Sierra Project-Team\\
 \'Ecole Normale Sup\'erieure\\
Paris, France \\
\texttt{francis.bach@inria.fr} \\
}

\begin{document}

\maketitle

\begin{abstract}
Structured sparsity has recently emerged in statistics, machine learning and signal processing as a promising paradigm for learning in high-dimensional settings. 
All existing methods for learning under the assumption of structured sparsity rely on prior knowledge on how to weight (or how to penalize) individual subsets of variables during the subset selection process, which is not available in general. Inferring group weights from data is a key open research problem in structured sparsity.
In this paper, we propose a Bayesian approach to the problem of group weight learning. We model the group weights as hyperparameters of heavy-tailed priors on groups of variables and derive an approximate inference scheme to infer these hyperparameters. We empirically show that we are able to recover the model hyperparameters when the data are generated from the model, 
and we demonstrate the utility of learning weights in synthetic and real denoising problems.
\end{abstract}

\section{Introduction}
\label{sec:intro}

High-dimensional prediction problems are more and more common in many application domains such as computational biology, signal processing, computer vision or natural language processing. To handle this high-dimensionality, one usually resorts to linear modeling and regularization with sparsity-inducing norms, such as the $\ell_1$ norm. This type of regularization results in \emph{sparse} models, meaning that the model is described by relatively few parameters. Besides making parameter learning consistent in high-dimensional settings, the sparsity assumption has the appealing property of yielding more interpretable models. As an example, consider the problem of explaining a particular phenotype of patients, e.g., the disease state, based on the genome sequence of each patient. Sparse linear approaches try to find a handful of genetic loci that govern the disease state, rather than a model involving the whole sequence. The $\ell_1$-regularized sparse linear models, such as the LASSO \citep{Tibshirani94} or basis pursuit \citep{chen}, are well studied by now, with a solid body of theoretical results, efficient algorithms and applications in diverse fields \citep[see, e.g.,][and references therein]{BuhGee11}. However, in practice, we often know that there is more \emph{structure} in the problem at hand, which cannot be captured by simple sparse modeling and $\ell_1$ regularization, and which, if exploited, can improve the estimation of parameters as well as the interpretability of the estimates \citep[see][and references therein]{Cevher2008,Huang2011,Bachetal12a}.
In our example, we could expect the genetic loci that influence the disease to be part of a small number of connected patterns in a known gene-gene interaction network \citep{Rapetal07,Azencottetal13}. In other words, we could be looking for a small number of possibly overlapping subsets of variables such that each subset corresponds to a connected subgraph in a given gene network, and the combination of variables in each subset influences the phenotype.

Given prior knowledge about the relevance of each considered group of variables, several methods exist for learning sparse models guided by this prior knowledge. 
These methods achieve different kinds of structured sparsity by regularization (penalization, weighting) with appropriate sparsity-inducing norms, that often correspond to convex relaxations of combinatorial penalties on the support (i.e., the set of indices of non-zero components) of the parameter vector. After the group LASSO \citep{YuaLin06}, a number of convex penalties have been proposed, generalizing the group LASSO penalty to the cases of overlapping groups \citep{ZhaYu09, JacOboVer09, JenAudBac11, Chenetal12}, including tree-structured groups \citep{KimXin10,Jenattonetal11}. See \citep{Bachetal12,Bachetal12a} for a more detailed review of sparsity-inducing norms.

\begin{figure}[t]
\begin{center}
\includegraphics[width=.5\textwidth]{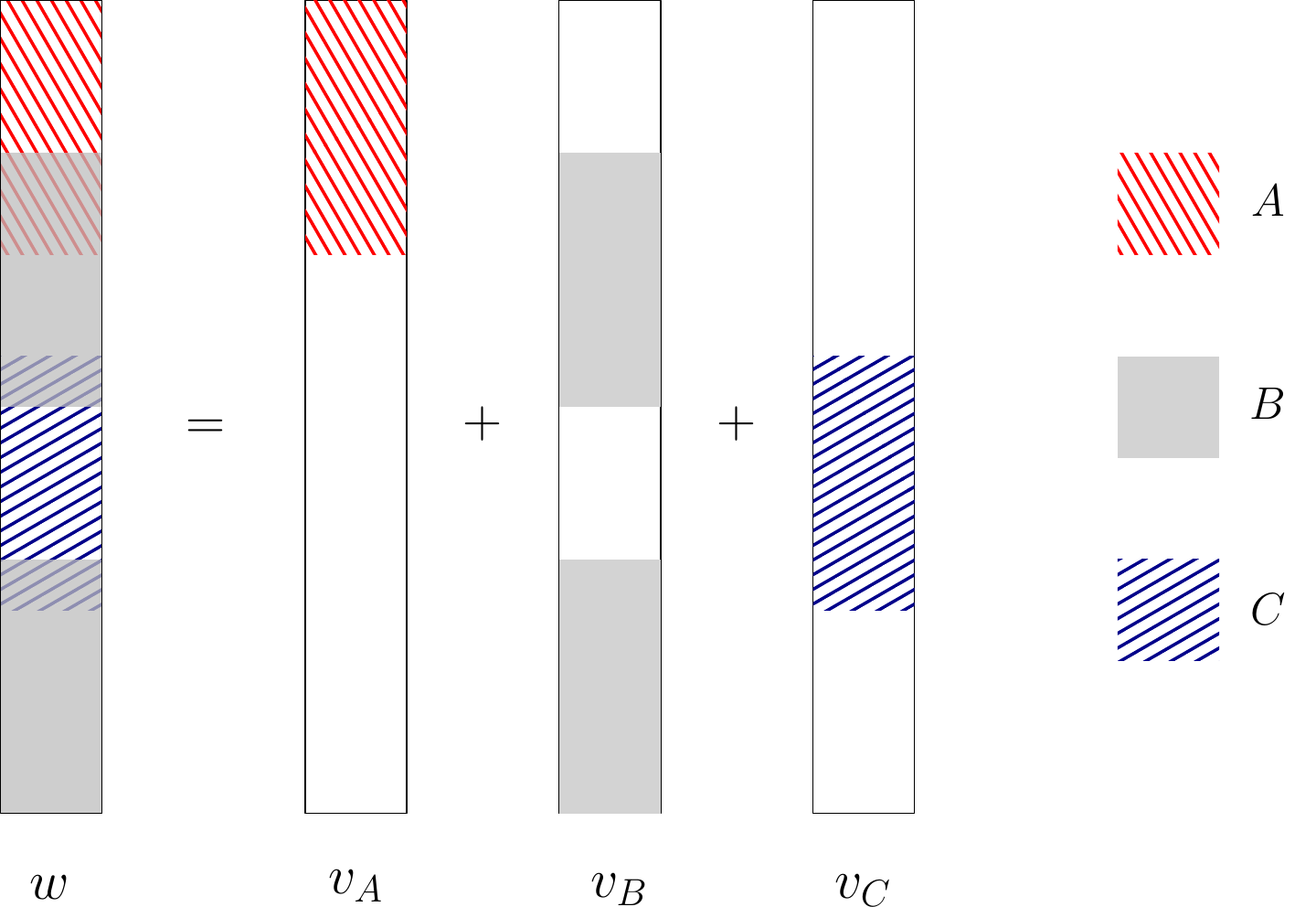}

\vspace{-.2cm}

\caption{The coefficient vector $w$ is covered by latent variables supported on subsets $A$, $B$ and $C$: $w = v_A+v_B+v_C$.}
\label{fig:w}
\end{center}
\end{figure}

While most of these norms induce \emph{intersection-closed} sets of non-zero patterns, \citet{JacOboVer09} and \citet{OboBac12} introduce a different, latent formulation of sparsity-inducing norms that yields \emph{union-closed} sets of non-zero patterns, meaning that the parameter vector $w$ is represented as a sum of latent vectors $v_A$, identically zero at indices not in~ $A$ for a subset $A$ of indices. If several such sets of indices are considered, then the support of $w$ (i.e., the set of indices $i$ for which $w_i$ is non-zero) is included in the union of such sets (see Figure~\ref{fig:w} for illustration with three sets $A$, $B$ and $C$).

In order to quantify the intuition above, \citet{OboBac12} consider the following function on the support ${\rm supp}(w)$ of $w$:
\begin{equation}
g({\rm supp}(w)) = \min_{\substack{\Acal' \subseteq \Acal,\\ \cup_{A\in\Acal'} A ={\rm supp}(w)}} \sum_{A\in\Acal'} f(A),
\end{equation}
that is, $g({\rm supp}(w))$ is the minimum-weight \emph{cover} of ${\rm supp}(w)$ with the subsets $A$ in the family $\Acal$. The weights $f(A)$ express our prior belief in the subset~$A$ being relevant: If a group $A$ is irrelevant, then $f(A)=\infty$. Using the function $g$ as a regularizer (essentially the approach of~\citet{Huang2011}) will encourage the support of the parameter vector $w$ to be a union of subsets $A \in \Acal$ with finite $f(A)$.

Moreover, \citet{OboBac12} computed a convex relaxation of the function $g$ defined above, leading to the following norm~$\Omega(w)$ equal to:
\begin{equation}
\label{eq:norm_guillaume}
  \min_{v_A\in\RR^P} \!\sum_{A\in\Acal} \!\|v_A\|_2 f(A)^{1/2}  {\rm \quad s.t.}   \sum_{A\in\Acal} v_A\!=\!w.  
\end{equation}
However, generally we do not have this prior knowledge about the relevance of individual groups: The problem of automatically choosing appropriate weights for groups of variables, $f(A)$, is an important open research problem in structured sparsity. Assuming that we have several learning problems with similar structure (the relevance of a given group is largely shared across individual problems), in this paper we propose a framework for learning group relevances from data. Note that learning the structure is naturally a multi-task problem, as it is impossible to estimate the prior on a vector of parameters if we only observe one particular instance of it.
To come back to our example, we could assume that we have several phenotypes that can be explained by groups of loci whose relevance is largely shared across phenotypes.
A recent approach to learning group relevances from data has been proposed by \citet{HerHer13}. However, this work only considers learning relevances of pairs of variables and does not make the link with sparsity-inducing norms. Let us also mention that probabilistic modeling for structured sparsity has also been explored by \citet{MarMur09} and \citet{MarSchMur09} in the context of learning Gaussian graphical models, and by \citet{Hanetal14} for multi-task learning with structure on tasks.

We approach the problem using probabilistic modeling with a broad family of heavy-tailed priors and derive a variational inference scheme to learn the parameters of these priors. Our model follows the pattern of \emph{sparse Bayesian} models \citep[][among others]{Palmeretal06,SeeNic11}, that we take two steps further: First, we propose a more general formulation, suitable for structured sparsity with any family of groups; Second, we learn the prior parameters from data. 
We show that prior parameter estimation with classical variational inference does not always lead to reasonable estimates in these models, and find a way of regularizing that works well in practice. Moreover, we propose a greedy algorithm that makes this inference scalable to settings in which the number of groups to consider is large.  In our experiments, we show that we are able to recover the model parameters when the data are generated from the model, and we demonstrate the utility of learning penalties in image denoising.

\section{A Probabilistic Model for Structured Sparse Linear Regression}
\label{sec:model}
In this section, we formally describe our model and a suitable approximate inference scheme. 

\subsection{Model definition}
We consider $K$ linear regression problems with design matrices $X^k \in \RR^{N^k\times P}$ and response vectors $y^k \in \RR^{N^k}$ for $k\in\{1,\ldots, K\}$. For each~$X^k$ and $y^k$, we assume the classical Gaussian linear model with i.i.d.~noise with variance~$\sigma^2$, that is, 
\begin{equation}
\label{eq:distr_y}
y^k \sim \Ncal(X^kw^k, \sigma^2 I).
\end{equation}

Let $V$ be the set of indices of variables $\{1,\ldots,P\}$. For a family $\Acal$ of subsets of $V$, we assume 
\begin{equation}
\label{eq:w}
\displaystyle w^k = \sum_{A\in \Acal} v_A^k,
\end{equation}
where, for each $k$, 
\begin{itemize}
\item $\forall A\in\Acal, v_A^k$ is a vector in $\RR^P$ such that all its components with indices in $V\setminus A$ are zero (in other words, it is supported on $A$), 
\item $\{v_A^k\}_{ A \in \Acal}$ are jointly independent, and 
\item $\forall A\in\Acal, v_A^k$ has an isotropic density with inverse scale parameter~$f(A)$
\begin{equation}
\label{eq:prior_v_A}
p(v_A^k|f(A))=q_A(\|v_A^k\|_2 f(A)^{1/2})f(A)^{|A|/2},
\end{equation}
where $q_A$ is a heavy-tailed distribution that only depends on $A$ through its cardinality, $|A|$. We specify~$q_A$ in Section \ref{sec:super-Gaussian}.
\end{itemize}

We regard the inverse scale parameter~$f(A)$ as a measure of relevance of the group of variables~$A$\footnote{Abusing notation, we will call ``group $A$'' the subset of variables indexed by elements of $A$ throughout the paper.}: If a group of variables is irrelevant, then~$f(A)$ should equal infinity. 
We are interested in priors~$q_A$ such that for each task indexed by~$k$ only a handful of~$v_A^k$ can be significantly away from~zero. 

Here it is important to stress the link between the expression of our isotropic prior \eqref{eq:prior_v_A} and the norm~$\Omega(w)$ \eqref{eq:norm_guillaume} from \citet{OboBac12}, introduced above: 
The log-likelihood of parameter vectors $\{w^k\}_{k=1,\ldots,K}$ with respect to $f$ will (up to a constant) be equal to the term $\sum_{A\in\Acal} \log q_A(\|v_A^k\|_2 f(A)^{1/2})$, which very closely resembles the norm \eqref{eq:norm_guillaume}. If~$q_A$ is the generalized Gaussian distribution (cf. Section \ref{sec:specialcases}), the two expressions match exactly. Thus, learning with our prior is a natural probabilistic counterpart of learning with the sparsity-inducing norm~\eqref{eq:norm_guillaume}.

Given data $\{X^k, y^k\}_{k=1,\ldots,K}$ and such a model for the prior, our goal will be to infer the parameters~$f(A)$ by maximizing the likelihood with respect to~$f$,
\begin{equation}
\label{eq:ML_brute}
\begin{aligned}
\ p(y^1,\ldots,y^K |f) = \prod_{k=1}^K \int p(y^k|X^kw^k,\sigma^2I) \prod_{A\in \Acal}p(v_A^k|f(A))  dv_A^k,\\
\end{aligned}
\end{equation}
where the parameters~$v_A^k$ are marginalized.

\subsection{Super-Gaussian priors}
\label{sec:super-Gaussian}
We assume that $q_A$ is a \emph{scale mixture of Gaussians}, i.e.,
\begin{equation*}
q_A(u) = \int_0^{\infty} \Ncal(u|0,s) r_A(s) ds
\end{equation*}
for some mixing density $r_A(s)$. 
The main reason why we choose to work with the family of scale mixtures of zero-mean Gaussians is that it contains distributions that are heavy-tailed and therefore suitable for modeling sparsity; One such distribution is Student's $t$ which we use in our experiments.
The inverse scale parameter of the distribution on $v_A^k$,  $f(A)$, captures the relevance of the group~$A$: the smaller $f(A)$, the more relevant the group, that is, the larger the values $v_A^k$ is likely to take. Note that even if the group $A$ is relevant, not all $v_A^k, k=1,\ldots,K$ have to be large. In fact, if the parameters $v_A^k, k=1,\ldots,K$ are drawn from a heavy-tailed distribution with small $f(A)$, then only a fraction of them will be significantly away from zero. Moreover, as we show in Section \ref{sec:variational}, learning in such models is amenable to variational optimization with closed-form updates and leads to an approximate Gaussian posterior on~$v_A^k$. 

In general, the integral in~\eqref{eq:ML_brute} is intractable for Gaussian scale mixtures, therefore one has to resort to sampling or approximate inference to learn parameters in such models.
The fact that $q_A$ is a Gaussian scale mixture implies that it is also \emph{super-Gaussian}, that is, the logarithm of $q_A(u)$ is convex in $u^2$ and non-increasing \citep{Palmeretal06}\footnote{Note that the converse is not true: complete monotonicity of the log-density is a necessary and sufficient condition for the existence of a Gaussian scale mixture representation~\cite[Section 3]{Palmeretal06}.}. It therefore admits a representation of the following form by convex conjugacy
\begin{equation}
\label{eq:q_superG}
\log q_A(u) = \sup_{s \geq 0} -\frac{u^2}{2s} - \phi_A(s),
\end{equation}
where $\phi_A(s)$ is convex in $1/s$. Note that the expression inside the supremum in \eqref{eq:q_superG} has a unique maximizer. 
In this work we only consider~$q_A$ for which this maximizer has an analytical simple form. From~\eqref{eq:prior_v_A} and~\eqref{eq:q_superG}, we get the following variational representation for $p(v_A^k|f(A))$: 
\begin{equation}
\label{eq:var_repr_p_v_A}
\begin{aligned}
p(v_A^k|f(A))& = f(A)^{\frac{|A|}{2}} \!\! \sup_{\zeta_A^k \geq 0} \exp{\Big( -\frac{\|v_A^k\|_2^2f(A)}{2\zeta_A^k} - \phi_A(\zeta_A^k) \Big)}\\
& = f(A)^{\frac{|A|}{2}} \!\! \sup_{\zeta_A^k \geq 0} \! \Big[ \! \Ncal \! \Big(v_A^k \Big| 0,\frac{\zeta_A^k I}{f(A)} \Big) \!\Big(2\pi\frac{\zeta_A^k}{f(A)}\Big)^{\!\!\!\frac{|A|}{2}} \! \! \! e^{-\phi_A(\zeta_A^k)} \Big].
\end{aligned}
\end{equation}
For a particular choice of the prior $q_A$, we measure the relevance of the group of variables $A$ by the expectation of $\|v_A^k\|_2^2$ (which amounts to the sum of the variances of the individual components of~$v_A^k$), 
\begin{equation*}
\EE\big[\|v_A^k\|_2^2\big] = \frac{\EE_{\|z\|_2\sim q_A} \big[\|z\|_2^2\big]}{f(A)},
\end{equation*}
where $\EE_{\|z\|_2\sim q_A} \big[\|z\|_2^2\big]$ is the expectation of $\|z\|_2^2$ under the standardized distribution $q_A$ on $\|z\|_2$. In fact, as we have
\begin{equation*}
\EE\big[\|w^k\|_2^2 \big] = \sum_{A\in\Acal} \EE\big[\|v_A^k\|_2^2 \big]
\end{equation*}
given our independence assumption, the expected value of $\|v_A^k\|_2^2$ allows us to measure the contribution of the group $A$ with respect to $\EE\big[\|w^k\|_2^2 \big]$. We somewhat abusively call $\EE\big[\|w^k\|_2^2 \big]$ the \emph{signal variance} in our experiments, as opposed to $P\sigma^2$, the \emph{noise variance}.
\begin{figure}
\begin{center}
\includegraphics[width=.47\textwidth]{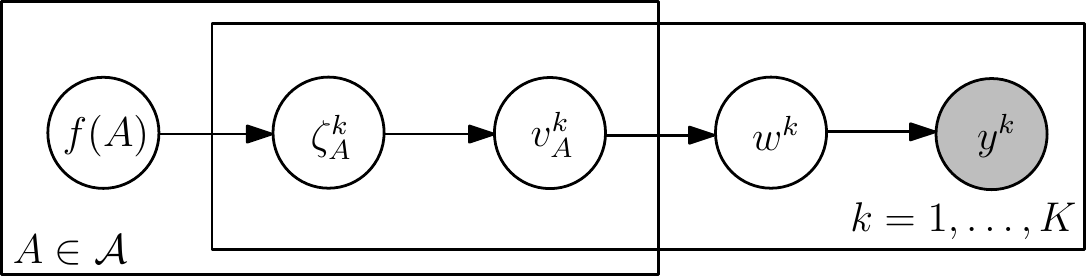}
\caption{The graphical representation of our model.}
\label{fig:graphical_model}
\end{center}
\end{figure}
Figure~\ref{fig:graphical_model} represents the graphical model corresponding to our assumptions. 
Note that we have explicitly incorporated the variational parameter $\zeta_A^k$ into the graphical model: In fact, the same parameter can also be interpreted as the scale parameter of the Gaussian in the Gaussian scale mixture representation of $p(v_A^k|f(A))$ \citep{Palmeretal06}.

\subsection{Inference}
\label{sec:variational}
Our model described above, namely the combination of the density of $y^k$ \eqref{eq:distr_y} and the variational representation of the prior density on~$v_A^k$~\eqref{eq:var_repr_p_v_A}, leads to the following variational bound on the marginal distribution of~$y^k$:
\begin{equation*}
\begin{aligned}
\log &\,p(y^k|f)\\
& =\log \int p(y^k|X^kw^k,\sigma^2I) \prod_{A\in \Acal}p(v_A^k|f(A))  dv_A^k\\
&\geq \sup_{\substack{{\zeta_A^k \geq 0}\\{ A \in \Acal}}}  \Big\{\log \Ncal(y^k|0, X^kM  Z^kF^{-1} M^{\top}{X^k}^{\top} + \sigma^2I) \\[-.1cm]
& + \sum_{A\in\Acal} \Big[ \frac{|A|}{2}\log f(A) \! + \! \frac{|A|}{2}\log\Big(2\pi\frac{\zeta_A^k}{f(A)}\Big) \!-\! \phi_A(\zeta_A^k)  \Big]\Big\},\\
\end{aligned}
\end{equation*}
where $M$ is a matrix of dimension $P\times\sum_{A \in \Acal} |A|$ that ensures $w^k = Mv^k$ where $v^k$ is the concatenation of all elements indexed by elements of $A$ in $v_A^k, A \in \Acal$, and $F$ and $Z^k$ are square diagonal matrices of size $\sum_{A \in \Acal} |A|$ whose diagonals consist of $f(A)$ and $\zeta_A^k$ respectively, replicated $|A|$ times, for each $A\in \Acal$.
Thus, as an approximation to minimizing the negative log-likelihood, 
we would like to minimize the following overall bound with respect to $f$ and~$\zeta_A^k$ for all $A \in \Acal$ and $k\in\{1,\ldots,K\}$:
\begin{equation}
\label{eq:LBgeneral_sum}
\begin{aligned}
-\sum_{k=1}^K \Big\{ &\!- \frac{1}{2} {y^k}^{\top} \!\Big(X^kM  Z^kF^{-1} M^{\top}{X^k}^{\top}\!\! + \sigma^2I\Big)^{-1} \! y^k \! - \!\frac{1}{2} \log\det\Big(X^kM  Z^kF^{-1} M^{\top}{X^k}^{\top}\!\! + \sigma^2I\Big) \\[-.1cm]
 & + \sum_{A\in\Acal}\frac{|A|}{2}\log f(A) + \frac{\sum_{A\in\Acal}|A|\!-\! N^k}{2}\log(2\pi) + \frac{1}{2}\log\det (Z^kF^{-1} ) -\sum_{A\in\Acal}\phi_A(\zeta_A^k)  \Big\}.\\
\end{aligned}
\end{equation}
In its form given by \eqref{eq:LBgeneral_sum}, the bound is difficult to optimize. However, we recognize parts of it as minima of convex functions, which allows us to design an iterative algorithm with analytic updates, finding a local minimum (see the appendix for details). Our optimization problem becomes 
\begin{equation}
\label{eq:obj}
\begin{aligned}
\inf_{\zeta^k\geq 0} \inf_{v^k} \inf_{\Sigma^k\succcurlyeq 0}  \sum_{k=1}^K\Big\{ &\frac{1}{2\sigma^2} \|y^k-X^kMv^k\|^2_2 + \frac{1}{2}  \sum_{A\in\Acal} \frac{f(A)}{\zeta_A^k} (\|v_A^k\|^2_2 + \tr \Sigma_{AA}^k) - \frac{1}{2} \log\det \Sigma^k\\[-.1cm] 
& + \frac{N^k}{2}\log(\sigma^2)  + \frac{N^k}{2} \log (2\pi)  + \frac{1}{2\sigma^2} \tr{M^{\top}{X^k}^{\top}X^kM\Sigma^k} -\frac{1}{2}  \sum_{A\in\Acal}|A|\\
&+  \sum_{A\in\Acal} \Big[ - \frac{1}{2}|A|\log f(A) -\frac{|A|}{2} \log 2\pi + \phi_A(\zeta_A^k) \Big]\Big\},\\
\end{aligned}
\end{equation}
and the closed-form updates are
\begin{equation}
\label{eq:updates}
\begin{aligned}
\Sigma^k & = \sigma^2(M^{\top}{X^k}^{\top}X^kM + \sigma^2 F{Z^k}^{-1} )^{-1}\\
v^k   & = (M^{\top}{X^k}^{\top}X^kM + \sigma^2 F{Z^k}^{-1})^{-1}M^{\top}{X^k}^{\top}y^k\\
\zeta_A^k & = \argmin_{z \geq 0} \phi_A(z) + \frac{1}{2} \frac{f(A)}{z} (\|v_A^k\|^2_2 + \tr \Sigma_{AA}^k)\\ 
\sigma^2& = \frac{\sum_{k=1}^K \!\! \big\{ \|y^k \!\!-\!\! X^kMv^k\|^2_2 \!+\!   \tr M^{\top}{X^k}^{\top}X^kM\Sigma^k \big\}} {\sum_{k=1}^{K}N^k} \\
 f(A) & = \frac{K|A|}{\sum_{k=1}^K \frac{1}{\zeta_A^k} (\|v_A^k\|^2_2 + \tr \Sigma_{AA}^k)},\\
\end{aligned}
\end{equation}
iterated until convergence. 
\begin{remark}
Note that the only update that depends on the specific prior distribution is that for the variational parameter $\zeta_A^k$, all others apply to all super-Gaussian priors. 
\end{remark}
\begin{remark}
It can be shown that the updates \eqref{eq:updates} exactly correspond to the updates yielded by mean-field variational inference in the special case of Gaussian scale mixtures \citep{Palmeretal06}. However, the approach presented here is more general, as it also applies to super-Gaussian priors that are not Gaussian scale mixtures.
\end{remark}
\begin{remark}
Using the matrix inversion lemma, the update for $\Sigma^k$ can be rewritten in such a way that we avoid the expensive inversion of a $\sum_{A\in\Acal}|A|\times \sum_{A\in\Acal}|A|$ matrix and we only have to invert a $P\times P$ or $N^k\times N^k$ matrix instead, which can even be diagonal in certain cases (see the appendix for details). 
When it is not diagonal, matrix inversions can be avoided by making an extra diagonal assumption on the covariance matrix of the Gaussian posteriors of all $v_A^k$.
\end{remark}
\begin{remark}
While we do provide an update equation for $\sigma^2$ for completeness, in general it is customary to assume the noise level known, which we also do in all our experiments. 
\end{remark}

\subsection{Special cases}
\label{sec:specialcases}
The family of super-Gaussian distributions includes Student's $t$ and generalized Gaussian distributions among many others. We here give the densities of these distributions, as well as the expressions for the quantities in our model and inference that depend on the particular prior on~$v_A^k$.

\paragraph{Student's $t$:} The density of this distribution is given by 
\begin{equation}
 p(v_A^k|a, f(A)) =f(A)^{\frac{|A|}{2}} \frac{\Gamma( a + |A|/2)}{\Gamma(a)} \Big(\frac{1}{2\pi}\Big)^{\frac{|A|}{2}}\Big(1 + \frac{{\|v_A^k\|_2}^2f(A)}{2} \Big)^{-a-\frac{|A|}{2}},
\end{equation}
where $a$ is a parameter governing the shape of the distribution. The smaller $a$, the heavier-tailed the distribution (for $a\le 1$, there is no finite variance).
For this distribution, 
\begin{equation}
\begin{aligned}
\phi_A(\zeta_A^k)   = & \frac{1}{\zeta_A^k}+ (a+1/2)\log(\zeta_A^k) + \frac{|A|}{2}\log(2\pi)  - (a + |A|/2)+(a+|A|/2)\log(a+|A|/2) \\
              &   - \log(\Gamma(a+|A|/2))+ \log(\Gamma(a)),\\
\end{aligned}
\end{equation}
and, therefore, the update for $\zeta_A^k$ is written as
\begin{equation}
\zeta_A^k   =  \frac{1+\frac{1}{2} f(A) (\|v_A^k\|^2_2 + \tr \Sigma_{AA}^k)}{a+\frac{|A|}{2}} .
\end{equation}
The variance of a Student's $t$-distributed random variable, if $a>1$, is
$\EE(v_A^k {v_A^k}^{\top})=\frac{1}{f(A)(a-1)}I,
$
and therefore 
$\EE(\| v_A^k\|_2^2)=\frac{|A|}{f(A)(a-1)}.
$
Student's $t$ has a natural representation as a Gaussian scale mixture with the inverse Gamma as the mixing distribution. 
All our experiments are carried out using Student's~$t$.

\paragraph{Generalized Gaussian:} The density is given by
\begin{equation}
 p(v_A^k|\gamma, f(A)) = f(A)^{\frac{|A|}{2}}\frac{\frac{\gamma}{2}\Gamma(\frac{|A|}{2})} {\pi^{\frac{|A|}{2}} \Gamma(\frac{|A|}{2} )} e^{-\|v_A^k f(A)^{\frac{1}{2}}\|_2^\gamma } 
\end{equation}
\citep{Pascaletal13}. Consequently, we have 
\begin{equation}
\begin{aligned}
\phi_A(\zeta_A^k)   = & -\log \frac{\frac{\gamma}{2}\Gamma(\frac{|A|}{2})} {\pi^{\frac{|A|}{2}} \Gamma(\frac{|A|}{2} )} + \frac{{\zeta_A^k}^\frac{\gamma}{2-\gamma}( \frac{1}{\gamma} - \frac{1}{2}) }{\gamma^{\frac{2}{\gamma-2}} },\\
\end{aligned}
\end{equation}

\begin{equation}
\zeta_A^k   =  \Big(-\frac{\frac{1}{2} f(A) (\|v_A^k\|^2_2 + \tr \Sigma_{AA}^k)}{(1/\gamma -1/2)\gamma^{\frac {\gamma-2}{2}}  }  \frac{\gamma-2}{\gamma} \Big)^{\frac{2-\gamma}{2}},
\end{equation}
and $\EE(\| v_A^k\|_2^2)=\frac{\Gamma(|A|/\gamma + 2/\gamma)}{f(A)\Gamma(|A|/\gamma)}$.

\subsection{Learning with all groups}
\label{sec:allgroups}
While our model and the associated inference algorithm described earlier are valid for any set of groups~$\Acal$, including $\Acal=2^V$, the algorithm is impractical when $\Acal$ is large: Indeed, even if we only have 20 variables and 1000 tasks, learning with $\Acal = 2^{\{1,\ldots,20\}}$ implies that the number of variational parameters~$\zeta_A^k$ will exceed a billion. To avoid working with a prohibitively large number of groups at once, one can leverage an \textit{active set}-type heuristic that maintains a list of relevant groups and iteratively updates it. Algorithm~\ref{alg:greedy}, which we discuss in detail in the following, describes one way to do this. It requires setting the maximal allowed cardinality $T$ of $\Acal$, and the number $D$ of groups to be discarded in each active set update. We start by learning with singletons only (steps 1 and 2); After ranking the groups in $\Acal$ according to their relevance measured by $\frac{f(A)}{|A|}$ into the sequence $(A_1,\ldots,A_{|\Acal|})$ (step 3), we determine the additional groups to be considered, $\Acal'$, by taking the first $T$ sets from the sequence $(A_1\cup A_2 ,\ldots, A_1 \cup A_{|\Acal|}, A_2\cup A_3,\ldots, A_2\cup A_{|\Acal|}, \ldots)$, ignoring groups that have been considered in the past and making sure we do not add the same group more than once; In steps 5-11 we repeatedly (a) learn with $\Acal\cup\Acal'$, (b) rank the groups, (c) update $\Acal$ and $\Acal'$. In step 8 we choose not to discard the singletons just to make sure that $\Acal$ always covers $\{1,\ldots,P\}$. The stopping criterion (step 5) may be that we have no more groups to consider (if $P$ is small enough), or that we have reached a predefined maximal number of iterations.
\begin{algorithm}
\caption{Active set procedure for the discovery of relevant groups}\label{alg:greedy}
\begin{algorithmic}[1]
\REQUIRE $T\in\NN$, $D\in\NN$
\STATE Let $\Acal =\{1,\ldots,P\}$ and $\Dcal=\emptyset$
\STATE $f \gets \text{variational}(\Acal)$
\STATE Rank all $A\in\Acal$ according to their relevance
\STATE Determine $\Acal'$ \\
(make sure $\left\vert\Acal\cup\Acal'\right\vert\!\le\! T, \Acal'\cap(\Acal\cup\Dcal)\!=\!\emptyset$)
\WHILE {stopping condition not met}
\STATE $f \gets \text{variational}(\Acal \cup \Acal')$
\STATE Rank $A\!\in\Acal\! \cup \!\Acal'$ according to their relevance
\STATE Add to $\Dcal$ the $D$ least relevant non-singletons in $\Acal\cup\Acal'$ 
\STATE $\Acal \gets \Acal\cup\Acal' \setminus\Dcal$
\STATE Determine $\Acal'$\\
(make sure $\left\vert\Acal\cup\Acal'\right\vert\!\le\! T, \Acal'\cap(\Acal\cup\Dcal)\!=\!\emptyset$)
\ENDWHILE
\end{algorithmic}
\end{algorithm}

\section{Approximation Quality and Regularization}
\label{sec:reg}

The goal of this section is to experimentally study the behavior of our approximate inference scheme in terms of estimation quality and to clarify how we can control it.
As we empirically show below, the variational approximation scheme from Section \ref{sec:variational} tends to overestimate the variance of the prior distribution (i.e., underestimate the inverse scale parameter $f(A)$) when this variance is smaller than~$\sigma^2$, the noise variance. This is undesirable, as we would like $f(A)$ to tend to infinity for irrelevant groups of variables. 
To circumvent this problem, we use an improper hyperprior of the form $p(f(A)) \propto f(A)^\beta$ to encourage $f(A)$ to go to infinity when the variance of $p(v_A)$ is smaller than $\sigma^2$. 
Consequently, the regularization term $-K\beta\sum_{A\in\Acal}\log f(A)$ with $\beta>0$ is added to the objective function \eqref{eq:obj}, and the only update that changes is that for $f(A)$:
\begin{equation}
 f(A) = \frac{K(\beta + \frac{|A|}{2})}{\frac{1}{2}\sum_{k=1}^K \frac{1}{\zeta_A^k} (\|v_A^k\|^2_2 + \tr \Sigma_{AA}^k)}.
\end{equation}
Thus, we substitute the approximate type-II maximum likelihood estimation of $f(A)$ by approximate (also ``type-II'') maximum a posteriori estimation. In Sections \ref{sec:exp_p_1} and \ref{sec:exp_p_2} we empirically study the effect of the parameter $\beta$ on the approximation quality.

\subsection{Scale parameter inference with only one variable}
\label{sec:exp_p_1}
In this experiment, we evaluate the performance of the variational method described in Section \ref{sec:variational} in recovering the unknown scale parameter $f$ of the prior in the simplest, 1-dimensional case (note that in this subsection we omit the subscripts $A$ as $\Acal =\{\{1\}\}$). More specifically, our goal here is to answer the following questions: Given an i.i.d. sample drawn from a univariate Student's $t$ with shape and inverse scale parameters $a$ and $f$, corrupted by Gaussian noise, and supposing we know both the noise variance $\sigma^2$ and the shape parameter $a$, can we precisely estimate the inverse scale parameter $f$ using the variational method from Section \ref{sec:variational}? In the settings where we cannot, does regularization improve our estimates?

\paragraph{Experimental setup.} We consider 10,000 tasks with one variable and one observation each ($P$, $N^k$ for all $k$, and $X^k$ for all $k$ equal to 1). Data are generated from the model with Student's $t$ prior on $v^k$ with parameters $a$ set to 1.5 and $f$ varying in the set $\mathcal{F}$ of 14 values between $0.02$ and $50$ taken roughly uniformly on the logarithmic scale, and Gaussian noise with variance $\sigma^2$ set to $1$. 
We compare the performance of the variational method with that of a grid search over $\mathcal{F}\cup\{10^5\}$, where we use the trapezoidal rule to numerically solve the intractable integral in \eqref{eq:ML_brute}. The grid search, feasible in this basic setting, provides the best available approximation to the regularized maximum likelihood solution.
To reduce the effect of random fluctuations, we repeat all experiments 5 times with different random seeds and report averaged results.

\begin{figure}

\begin{center}
\includegraphics[width=.95\textwidth]{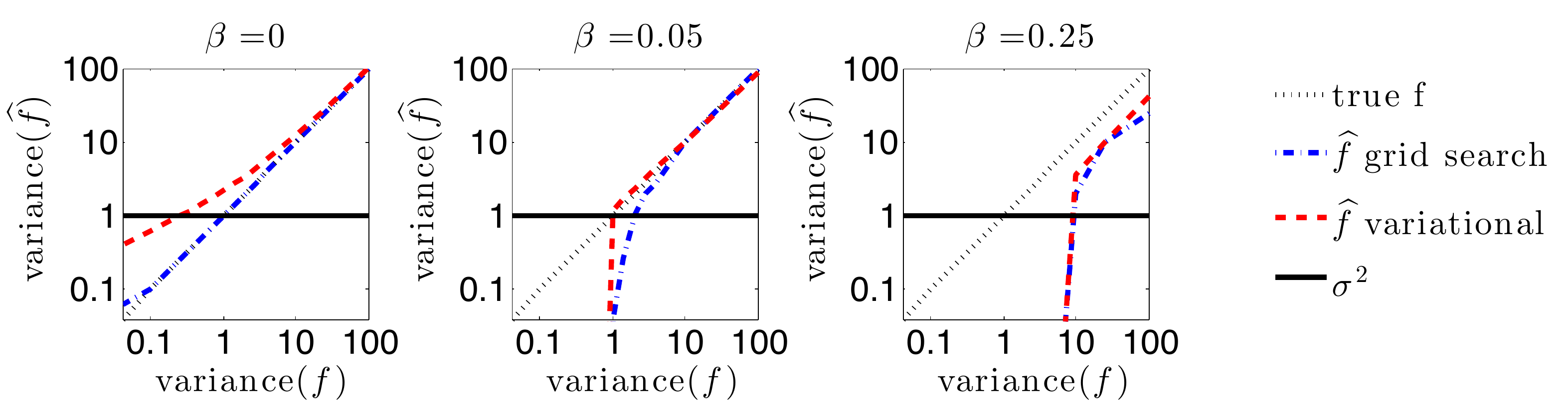}
\end{center}
\caption{Recovery of the variance of the univariate Student's $t$ distribution with added Gaussian noise of known variance with grid search and the variational method, with different levels of regularization. The x and y axes represent the variance based on the true and on the estimated $f$ parameter values, respectively.}

\label{fig:reg_scale_p1}
\end{figure}

\paragraph{Results.} Figure~\ref{fig:reg_scale_p1} summarizes the results. For three values of the parameter $\beta$, we plot (on the logarithmic scale) the estimated against the true variance for the considered range of the parameter $f$ (recall that the variance of a Student's $t$-distributed random variable with parameters $a$ and $f$ equals $\frac{1}{(a-1)f}$). In all figures, we also plot the variance of the Gaussian noise $\sigma^2$. We observe that in the absence of regularization ($\beta=0$) and when the signal is not much stronger than noise, the variational method overestimates the signal variance while the grid search does not. As we add regularization, this effect gradually goes away and the signal variance estimate is set to 0 (i.e., the estimate of $f$, $\widehat{f}$, goes to infinity) if the true signal variance is smaller than a certain threshold. When the regularization is too strong ($\beta=0.25$), the estimated signal variance drops to 0 even when the signal is stronger than the noise, and the variance of the signal is heavily underestimated. With the right amount of regularization ($\beta=0.05$ in this case) we observe the desired behavior: The variational method recovers the signal when it is stronger than noise, and sets $\widehat{f}$ to infinity otherwise. In all cases, variational estimates are close to the maximum likelihood estimates obtained by the grid search when the signal is much stronger than the noise.

\subsection{Structured sparsity with two variables}
\label{sec:exp_p_2}
In this section, we empirically study the most basic case of the group relevance learning problem. Suppose that in each task we only have 2 variables, and therefore 3 possible groups, $\Acal =\{\{1\},\{2\},\{1,2\}\}$. Let $X^k$ be the identity matrix in each task. In this basic setting, and supposing that the data come from the model, can our inference algorithm distinguish the case where the data $\{y^k\}_{k=1,\ldots,K}$ are generated by the group of variables $\{1,2\}$ from the opposite case, where the relevant groups are the two singletons $\{1\}$ and~$\{2\}$? 

\begin{figure}

\begin{center}
\includegraphics[width=.5\textwidth]{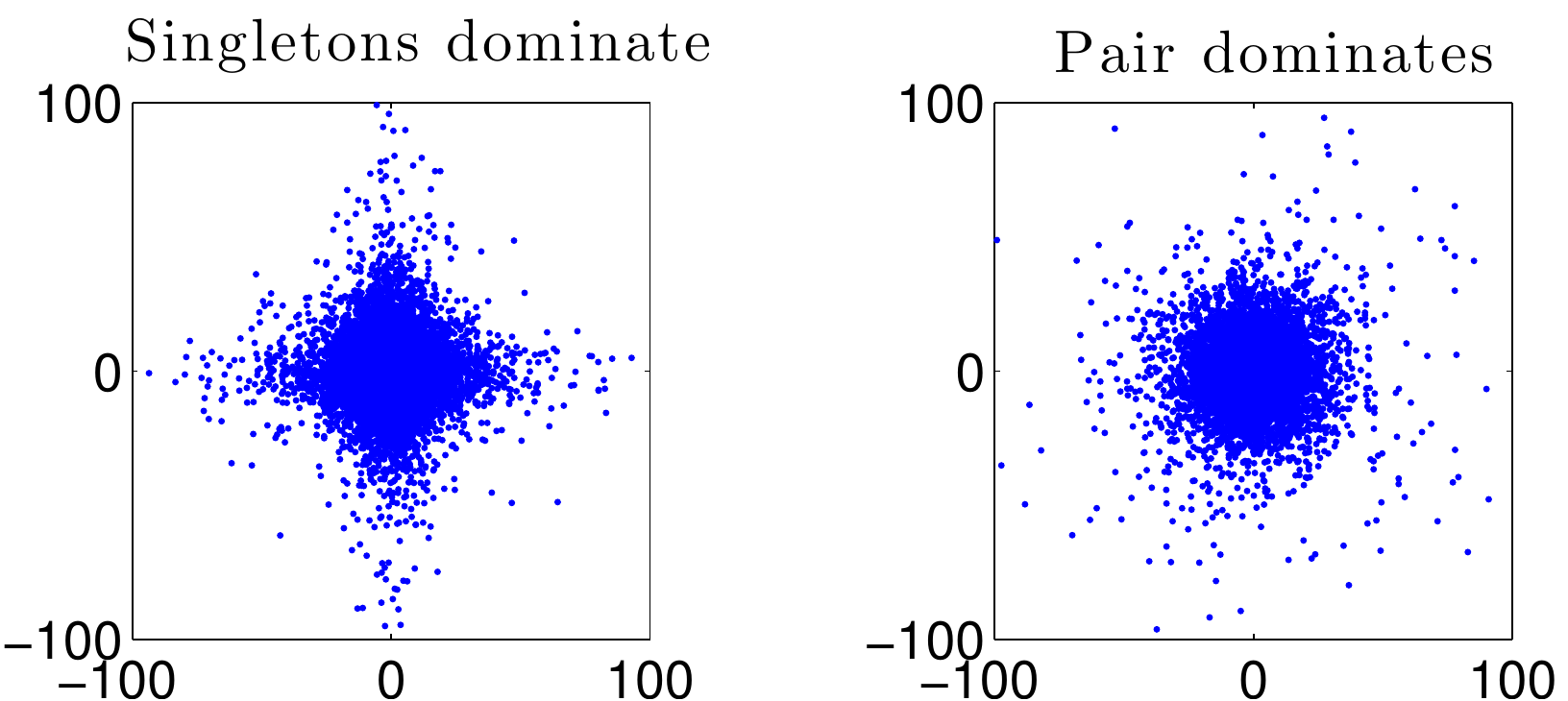}
\end{center}

\caption{On the left, the singletons are the relevant groups. On the right, the pair is the relevant group.}
\label{fig:singleton_pair_data}

\end{figure}

These two settings differ in fact significantly in the case of a heavy-tailed prior on $v_A^k$: We have $w^k = v_{ \{1\} }^k + v_{ \{2\} }^k + v_{ \{1,2\} }^k$; 
If $\{1, 2\}$ is relevant and $\{1\}$ and $\{2\}$ are not, 
then $v_{ \{1\} }^k$ and $v_{ \{2\} }^k$ will have to be close to zero for all $k$, however, $v_{ \{1,2\} }^k$ will be significantly far from zero for some $k$. As the prior on $v_A^k$ only depends on $v_A$ through its norm, these $v_{ \{1,2\} }^k$ can be anywhere on the circle with radius $\|v_{ \{1,2\} }^k\|_2$ with the same probability and therefore $y^k$ can also be anywhere on the circle with radius $\|y^k\|_2$. 
In contrast, 
when $\{1, 2\}$ is irrelevant and $\{1\}$ and $\{2\}$ are relevant, 
the rare events of $v_{ \{1\} }$ and $v_{ \{2\} }$ both being significantly away from zero will not occur at the same time for most $k$, and therefore the $y^k$ with a large norm will tend to be concentrated along the axes. This behavior (using Student's $t$ prior with parameter $a=1.5$ on $v_A^k$) is illustrated in Figure~\ref{fig:singleton_pair_data}, where we have plotted the data $\{y^k\}_{k=1,\ldots,K}$ for $K=5,000$ in both settings. 

\paragraph{Experimental setup.} We consider 5,000 tasks with $P$ and $N^k$ for all $k$ equal to 2, with the set of groups $\Acal =\{\{1\},\{2\},\{1,2\}\}$. 
The data are generated from the model with Student's $t$ prior on $v^k$ with parameters $a$ set to 1.5 and each $f(A)$ varying in a set of 14 values between $0.01$ and $25$ taken roughly uniformly on the logarithmic scale ($f(\{1\})$ and $f(\{2\})$ always equal each other), and Gaussian noise with variance $\sigma^2$ set to $1$.

\begin{figure}[t]
\begin{center}
\includegraphics[width=.92\textwidth]{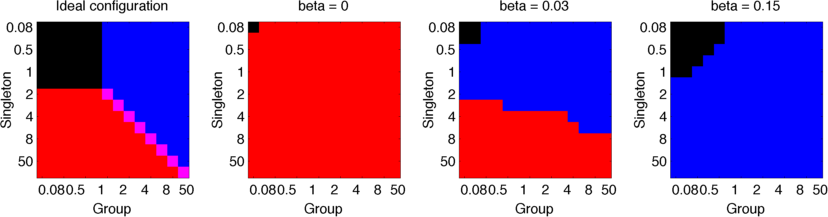}

\caption{A red (blue) square means that the estimate of the singleton (group) variance is larger than the estimate of the group (singleton) variance for the corresponding true singleton and pair variances indicated by the axes. A black square means that both singleton and pair variances are under $2\sigma^2$, the noise variance. Best seen in color.}
\label{fig:singleton_pair_results}
\end{center}
\end{figure}

\paragraph{Results.} Figure~\ref{fig:singleton_pair_results} summarizes the results for three values of the regularization parameter $\beta$ ($\beta=0$ corresponds to the absence of regularization). We report when the estimated pair variance $\frac{2}{(a-1)\widehat{f}(\{1,2\})}$ dominates (blue) or is dominated (red) by the estimated singleton variance $\frac{1}{(a-1)\widehat{f}(\{1\})}+\frac{1}{(a-1)\widehat{f}(\{2\})}$, provided that one of them is larger than the noise variance, $2\sigma^2$. We see that when we do not regularize, the variational method explains everything with the singletons. As we add regularization, the pair explains more and more variance, however in such a way that the pair also explains the signal coming from singletons. Nonetheless, there is a regime ($\beta=0.03$) where a strong signal coming from both the singletons and the pair is identified correctly. If we regularize too strongly ($\beta=0.15$), the entire signal is explained by the pair, regardless of its source.

\section{Experiments}
\label{sec:exp}

In our experiments we consider two different instances of the denoising problem and we empirically evaluate the performance of our approach in recovering both the signal and the structure. 

\subsection{Structured sparsity in the context of denoising}
In this section, we study toy multi-task structured sparse denoising problems. Our goal is to answer the following questions: Given data $\{y^k\}_{k=1,\ldots,K}$, generated from the model, and assuming that we know the true shape parameter $a$ of the Student's $t$ and the noise variance~$\sigma^2$, (a) can we recover the structure (i.e., the relevant groups and their weights), and (b) if we use the correct structure, is our denoising more accurate than when using a different structure?

\paragraph{Experimental setup.} To this end, we consider 10,000 tasks with $P$ and $N^k$ for all $k$ equal to 10, with the set of groups $\Acal =\{\{Q\}_{Q=1,\ldots,P}, \{1,\ldots,Q\}_{Q=2,\ldots,P}\}$. 
Each signal $w^k$ is generated using Student's $t$ with parameters $a$ set to $1.5$ and $f(A)$ set to 0.2 or to 200, depending on whether $A$ is considered relevant or irrelevant: In this fashion, the variance of the signal coming from relevant $A$ is $\frac{|A|}{(a-1)f(A)}=10\times|A|$ (respectively, $0.01\times|A|$ for irrelevant $A$). For each task~$k$, $y^k$ is a perturbed version of the signal $w^k$ with additive Gaussian noise of variance $\sigma^2I$.

We consider three different ways of generating data: 
\begin{itemize}
\item {\bf Singletons}: Here, only $\{1\},\ldots,\{5\}$ are relevant, all other groups in $\Acal$ are irrelevant. 
\item {\bf One group}: Only $\{1,2,3,4,5\}$ is relevant.
\item {\bf Overlapping groups}: The groups $\{1\}$, $\{1,2\}$,$\ \ldots\ $, $\{1,2,3,4,5\}$ are relevant. 
\end{itemize}
For the three cases, we choose $\sigma^2$ 
so that the total noise variance $P\sigma^2$ equals the total signal variance in each case.

We consider four models of increasing complexity for inference:
\begin{itemize}
\item {\bf LASSO-like}: In this simplest model, we only use the singletons, $\Acal = \{\{1\},\ldots,\{P\}\}$, and moreover, we force $f(A)$ to be constant across $\Acal$; In order to do so, we change the update for $f(A)$ to 
$f(A) = \frac{K\sum_{A\in\Acal}(\beta + \frac{|A|}{2})}{\frac{1}{2}\sum_{k=1}^K\sum_{A\in \Acal} \frac{1}{\zeta_A^k} (\|v_A^k\|^2_2 + \tr \Sigma_{AA}^k)}.$
This mimics the behavior of the LASSO, as the prior (that we are learning here) is the same for each coefficient.
\item {\bf Weighted LASSO-like}: The usual model with 
$\Acal \!=\! \{\{1\},\ldots,\{P\}\}$.
\item {\bf Structured}: The usual model with $\Acal \!=\!\{\{Q\}_{Q=1,\ldots,P},\{1,\!\ldots,\! Q\}_{Q=2,\ldots,P}\}$.
\item {\bf Structured (active set)}: The model where we also learn $\Acal$ using Algorithm~\ref{alg:greedy} (with parameters $T=4P$, $D=2P$, and 5 active set updates). 
\end{itemize}
We examine each of the 12 combinations of data generation and learning models. In each case, we use half of the tasks to find the optimal $\beta$ in terms of the mean squared prediction error (i.e., the mean squared difference between the true and the learned signals $w^k$) from a predefined range of 7 values, and the other half to learn with this $\beta$ and evaluate the test error.

\begin{table}[t]
\begin{center}
\begin{tabular}{|r|c| c |c|}
\hline
                        & Singletons                    & One group               & Overlapping  \\ \hline
LASSO-like       & 18.5$\pm$0.3             & 18.6$\pm$0.4         & 58.4$\pm$1.1\\  \hline
W. LASSO-like  & {\bf 14.5}$\pm$0.3     & 14.5$\pm$0.3         &  {\bf 42.8}$\pm$0.9\\  \hline
Structured        & 14.8$\pm$0.3              & {\bf 13.8}$\pm$0.3 &  43.0$\pm$0.9 \\  \hline
Structured(AS)  &  14.6$\pm$0.3      & 14.0$\pm$0.3 &  {\bf 42.8}$\pm$0.9 \\  \hline
\end{tabular}
\end{center}

\caption{Squared error averaged over the tasks with $95\%$-confidence error bars for each combination of data generation and learning models. The usage of boldface indicates that the corresponding method significantly outperforms the others, as measured using a $t$-test at the level $0.05$.\label{tab:12comb}}

\end{table}

\paragraph{Results.} We begin by examining the performance of each of the four models in \emph{signal recovery}: In Table \ref{tab:12comb} we report the mean squared error on the 5,000 test tasks with $95\%$-confidence error bars. For all three regimes for data generation, the LASSO-like model performs far worse than the three others in recovery. This is due to the fact that this model learns the same prior for all variables, although not all variables have the same marginal variance. In the first and third data generation regimes W.LASSO performs slightly better than Structured in signal recovery, while Structured has an advantage when a single group is relevant. The performance of Structured(AS) is systematically close to, or on a par with that of the best-performing model.

\begin{figure}[t]
\begin{center}
\includegraphics[width=.8\textwidth]{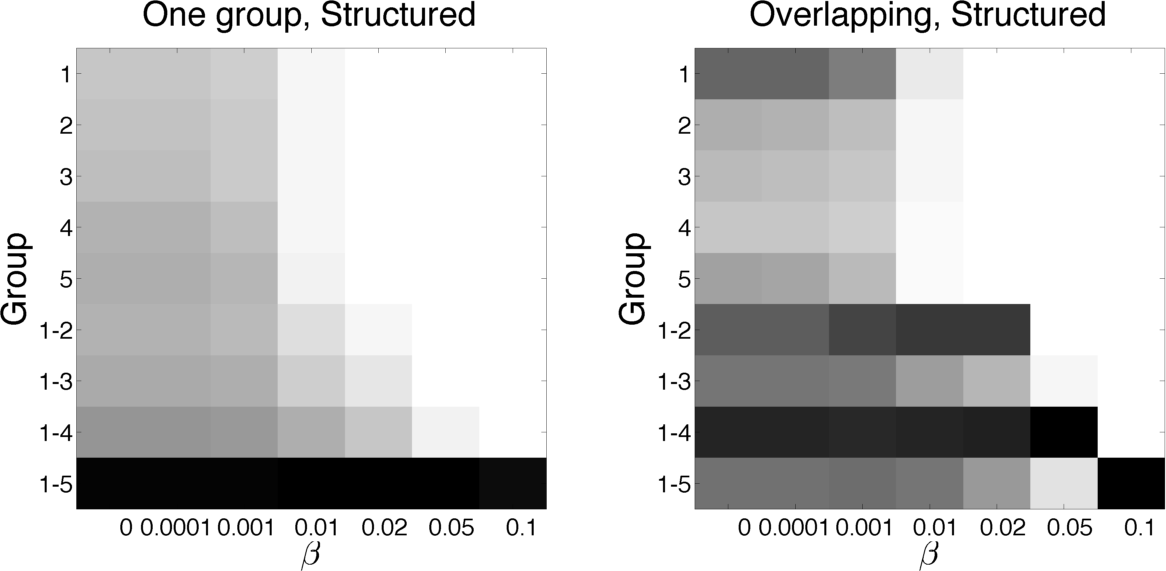}

\caption{For each group of variables on the y axis, the intensity of gray indicates the percentage of total explained variance per $\beta$. \label{fig:p10_exp_variance_23_3}}

\end{center}
\end{figure}

In terms of \emph{structure recovery}, for all three data generation regimes, we find one or more values of $\beta$ that lead to the recovery of the relevant groups by Structured and Structured(AS), with either the same or a slightly different $\beta$ value leading to the smallest error in signal recovery. Figure~\ref{fig:p10_exp_variance_23_3} illustrates the percentage of total explained signal variance by each group for the One group and Overlapping regimes and for the Structured model, for all considered regularization parameters: With no regularization, the model explains the signal with both the relevant group(s) and the singletons included in the relevant group(s), however with more and more regularization, the signal variance explained by smaller groups is taken over by larger ones. The groups containing elements from $\{6,\ldots,10\}$, not shown in the plot, explain no variance in no regularization regime, with the exception of the largest group $\{1,\ldots,10\}$ that explains the weak signal coming from the irrelevant groups (recall that we have non-zero signal variance $0.01\times|A|$ for the irrelevant groups $A$) in weak and moderate regularization regimes and takes over the whole signal variance when the regularization is too strong.

In summary, the performance in denoising does not change drastically depending on the amount of regularization, unless it is too strong; However, a small amount of regularization is likely to better capture the structure than no regularization; If there is a strong group structure among the variables, regularization may also lead to better recovery. 
A formal criterion to set the value of the hyperparameter $\beta$ would be to maximize its likelihood, as is customary in Bayesian methods.

\subsection{Image denoising with wavelets}
In this section, we consider the image denoising problem using wavelets. The Haar wavelet basis for 2-dimensional images~\citep{mallat} can naturally be arranged in three rooted directed quad-trees, which can be connected to form one tree by attaching the three roots to an artificial parent node; The structured sparsity-inducing norms with non-zero groups that are paths from the root in this tree have shown improvements over the $\ell_1$ norm~\citep{Jenattonetal11}. Our goal is to find out whether, in this task, (a) a value of $\beta$ that leads to good recovery for a set of images is also close to optimal for another set of images of roughly the same size, at least when the noise level is unchanged (stability of the hyperparameter); (b) learning a non-uniform prior on singletons improves recovery with respect to using a uniform prior (importance of learning a non-uniform prior); (c) learning the group structure helps beyond learning a non-uniform prior on singletons (importance of learning group relevances).

\begin{figure}[h]
\begin{center}
\includegraphics[width=0.2\textwidth]{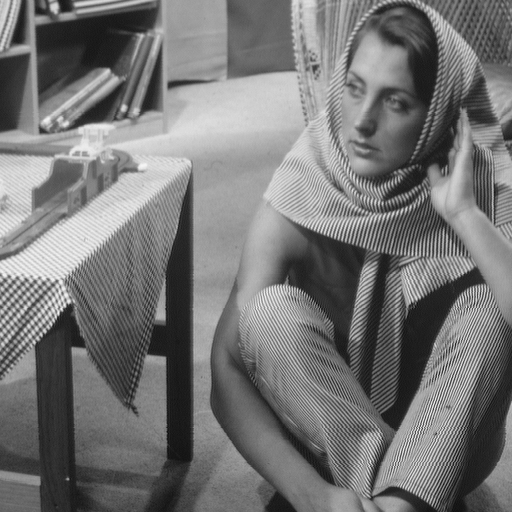} \quad \includegraphics[width=0.2\textwidth]{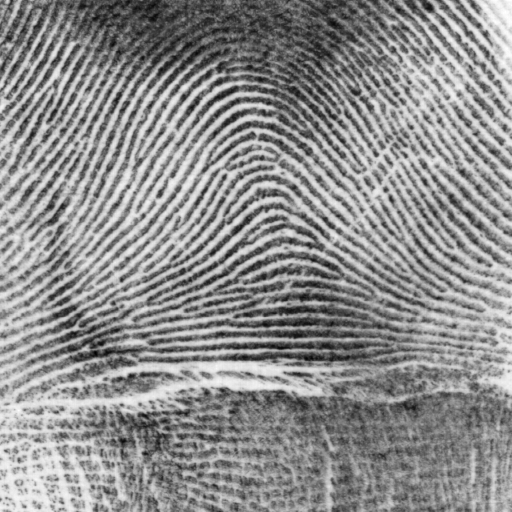}  \quad \includegraphics[width=0.2\textwidth]{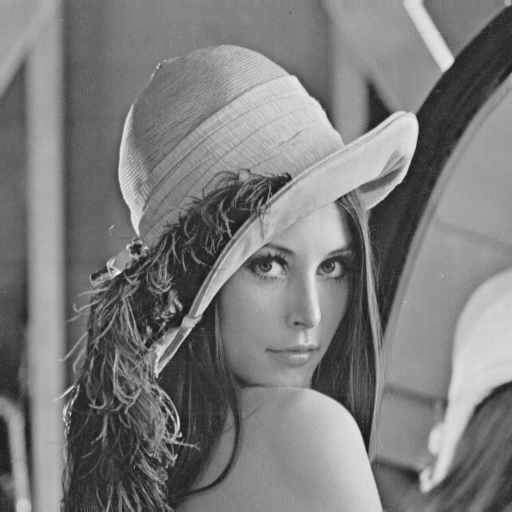} \quad \includegraphics[width=0.2\textwidth]{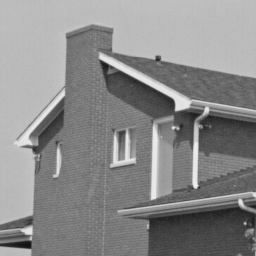}
\end{center}
\caption{The images used in our experiments (Barbara, Fingerprint, Lena, House).\label{fig:images}}
\vspace{.3cm}
\end{figure}

\paragraph{Experimental setup.} In order to denoise a large grayscale image, we cut it into possibly overlapping patches of $32\times 32$ pixels, which compose the multiple 1024-dimensional signals that we denoise simultaneously by learning the appropriate (structured) prior. We use four well-known images (see Figure~\ref{fig:images}), Barbara, Fingerprint, Lena ($512\times 512$ pixels each), and House ($256\times 256$ pixels). Each signal $w_k$ is formed by the wavelet coefficients of one $32\times 32$ patch. For each of the $K=961$ tasks ($841$ for House) we form $y^k$ by adding Gaussian noise of variance $\sigma^2=400$ along each dimension.
As in the previous section, we examine the performance of three instances of our model: the model with a uniform factorized sparse prior (LASSO-like), a non-uniform factorized sparse prior (W.LASSO-like), the structured norms on all descending (equivalently, ascending) paths in the rooted tree (Structured), and the structured norms on groups that we discover in the process of learning, with~2 active set updates (Structured(AS)). 
We consider a predefined range of 6 values for the regularization hyperparameter $\beta$, and 3 values ($0.5, 1.1, 1.5$) for the shape parameter~$a$ of Student's~$t$.
We compare the behavior of our methods with that of existing algorithms based on sparsity-inducing norms, which are not designed to learn group weights from data. From the family of such approaches, we choose the ``Tree-$\ell_2$'' structured norm proposed by \citet{Jenattonetal11}, and the classical LASSO \citep{Tibshirani94} on the wavelet coefficients. (We would like to stress here that ``Tree-$\ell_2$'' does need group weights to be specified, but does not provide a systematic way to learn them. They are usually set by introducing a group-weighting parameter $\alpha$ so that $\alpha^d$ is the weight of all groups at depth $d$ in the tree, and then optimizing $\alpha$ over a predefined range of values using cross-validation.)
We run these methods on each set of small images with the regularization parameter $\lambda$ and the group-weighting parameter $\alpha$ (only for Tree-$\ell_2$) varying over 
predefined ranges of 75 and 7 values 
respectively, and report the smallest error.
To train the LASSO and learn with the Tree-$\ell_2$ norm, we use the ``proximal'' toolbox of the software package SPAMS~\citep{Jenattonetal11}.

\paragraph{Results.} Table \ref{tab:mse_images_32} shows the best performance in terms of the mean squared error of each method on each image (which corresponds to a set of $K$ small images). The values in the parentheses for our proposed methods indicate the value of $\beta$ corresponding to the minimal error. The performance of our proposed methods with respect to the shape parameter $a$ is systematically slightly better for larger $a$, and all reported results correspond to $a=1.5$.  
According to these results, (a)~the performance of a given value of $\beta$ in signal recovery indeed seems to be stable across images (note that we have also observed that the performance on a given image is robust to small changes of the value of the hyperparameter); (b)~the fact that the LASSO and our LASSO-like model are systematically outperformed by models that weight each variable confirms the intuition that learning how to weight individual variables should boost the estimation quality; (c)~it seems that learning a prior on joint relevances of variables can lead to improved performance, as shown in the column corresponding to Fingerprint, although this is not always the case: on House and Lena, the performance of methods that learn group relevances is not significantly different from that of Tree-$\ell_2$, and in the case of Barbara they perform worse. Inspecting the relevances of different groups (paths in the wavelet tree) learned by Structured, we see that the groups explaining the bulk of the variance are overlapping groups of 2, 3, or 4 elements, mostly descending from the roots of the three quad-trees. In contrast, the relevant groups selected by Structured(AS) tend to consist of one to three roots of the three wavelet quad-trees and one or two wavelets of higher frequency, suggesting that paths in the wavelet tree may not always be the most natural groups in this problem. At last, let us stress that while ``Tree-$\ell_2$'' is applicable in problems where variables can be structured in a tree given in advance, our proposed approach applies to any known or unknown group structure.

The Matlab code used in our experiments is available at \url{http://cbio.ensmp.fr/~nshervashidze/code/LLSS}.

\begin{table} 
\begin{center}
\resizebox{0.98\linewidth}{!}{
\begin{tabular}{|r |c|c|c|c|}
\hline
        &     Barbara       & House & Fingerprint & Lena  \\ 
\hline
LASSO-like      & 179.0$\pm$4.6 (0.001)  & 107.5$\pm$2.6 (0.001)  & 247.5$\pm$1.7 (0.005)  & 110.3$\pm$2.8 (0.001)  \\ \hline
W.LASSO-like  & 163.3$\pm$5.1 (0)  & 93.7$\pm$2.6 (0)  & 195.0$\pm$1.8 (0.0001)  & 89.5$\pm$3.2 (0)  \\ \hline
Structured       & 164.8$\pm$5.3 (0)  & 95.3$\pm$2.9 (0)  & {\bf 193.6}$\pm$1.8 (0.0005)  & 90.3$\pm$3.5 (0)  \\ \hline
Structured(AS)  & 163.1$\pm$5.0 (0.0001) & 92.9$\pm$2.3 (0.0001) & 194.9$\pm$1.8 (0.001)  & 89.5$\pm$2.8 (0.0001) \\ \hline
 \hline
Tree-$\ell_2$&{\bf 155.3}$\pm$6.4 & 93.3$\pm$3.8 & 214.9$\pm$2.4 & 88.7$\pm$3.7 \\ \hline
LASSO  &176.7$\pm$6.4 & 102.1$\pm$3.6 & 250.0$\pm$2.2 & 106.6$\pm$3.9 \\ \hline
\end{tabular}
}
\end{center}
\caption{Squared error averaged over the images with $95\%$-confidence error bars for each combination of data generation and learning models. The usage of boldface indicates that the corresponding method significantly outperforms the others, as measured using a $t$-test at the level $0.05$. (Each number is divided by 1000 for readability.)\label{tab:mse_images_32}}
\end{table}

\section{Conclusions and Future Work}
\label{sec:concl}
In this paper, we have proposed a flexible and general probabilistic model and an associated inference scheme for automatically learning the weights of possibly overlapping groups in the context of structured sparse multi-task linear regression. We have shown that the classical variational inference scheme is not well adapted for learning with this model, and have proposed a regularization method that closes this gap. This has allowed us to investigate the effect of learning group weights in denoising problems, leading to the conclusion that learning penalties can significantly improve prediction quality, as well as the interpretability of the models, in this context. We have furthermore devised an active-set procedure that makes the inference with our model scalable to settings with large~$P$ and a large number of potential groups in~$\Acal$. 

In our future work we may consider different likelihood models to handle settings different from linear regression, such as binary classification. Learning group relevances for classification is indeed crucial, e.g., in the context of genome-wide association studies with binary phenotypes in computational biology, or for image segmentation in computer vision.

In the appendix we provide details on the derivation of the variational inference scheme for our model (briefly introduced in Section \ref{sec:variational}) and discuss efficient ways of implementing the closed-form updates~\eqref{eq:updates}.

\appendix 
{\setlength{\parindent}{0pt}
\setlength{\parskip}{2.2ex plus 0.5ex minus 0.2ex}

\section{Variational inference for the super-Gaussian structured sparse prior}
In this appendix, we derive step by step the variational updates given in Section 2.3.

We first recall our model: 
We assume that $\forall A\in\Acal, v_A^k$ has an isotropic density with inverse scale parameter $f(A)$
\begin{equation*}
\tag{\ref{eq:prior_v_A} revisited}
 p(v_A^k|f(A))=q_A(\|v_A^k\|_2 f(A)^{1/2})f(A)^{|A|/2},
\end{equation*}
where $q_A$ is a \emph{super-Gaussian} distribution, that is, the logarithm of $q_A(u)$ is convex in $u^2$ and non-increasing \citep{Palmeretal06}. It therefore admits a representation of the following form by convex conjugacy
\begin{equation*}
\tag{\ref{eq:q_superG} revisited}
\log q_A(u) = \sup_{s \geq 0} -\frac{u^2}{2s} - \phi_A(s),
\end{equation*}
where $\phi_A(s)$ is convex in $1/s$. Note that the expression under the supremum in \eqref{eq:q_superG} has a unique maximizer. 
In this work we only consider~$q_A$ for which this maximizer has an analytical simple form. From~\eqref{eq:prior_v_A} and~\eqref{eq:q_superG}, we get the following variational representation for $p(v_A^k|f(A))$: 
\begin{equation*}
\tag{\ref{eq:var_repr_p_v_A} revisited}
\begin{aligned}
p(v_A^k|f(A)) &= f(A)^{\frac{|A|}{2}} \sup_{\zeta_A^k \geq 0} e^{ -\frac{||v_A^k||^2f(A)}{2\zeta_A^k} - \phi_A(\zeta_A^k) }\\
& = f(A)^{\frac{|A|}{2}} \sup_{\zeta_A^k \geq 0} \Big[  \Ncal\Big(v_A^k| 0,\frac{\zeta_A^k}{f(A)} I\Big) \Big(2\pi\frac{\zeta_A^k}{f(A)}\Big)^{\frac{|A|}{2}} e^{-\phi_A(\zeta_A^k)} \Big].
\end{aligned}
\end{equation*}
Finally, as $v_A^k$ are assumed independent, 
\begin{equation*}
p(\{v_A^k\}_{A\in \Acal}|f) = \prod_{A\in\Acal} p(v_A^k| f(A)).
\end{equation*}

Our goal is to infer the set function $f$ from data by maximizing the type~II log-likelihood, $$\sum_{k=1}^K\log p(y^k|f).$$ We tackle the problem using variational inference and consider the following lower bound on $\log p(y^k|f)$ (obtained by combining the density of~$y^k$ \eqref{eq:distr_y}, the variational representation of the prior density on~$v_A^k$~\eqref{eq:var_repr_p_v_A}, and taking into account the independence of $\{v_A^k\}_{A\in\Acal}$), for sets $A \in \Acal \subseteq 2^V$ and for each regression task $k$, where we use the following notation: 
\begin{itemize}
\item $v^k$ is the concatenation of all elements indexed by elements of $A$ in $v_A^k, A \in \Acal$, 
\item $Z^k$ is a square diagonal matrix of dimension $\sum_{A \in \Acal} |A|$. Its diagonal consists of $\zeta_A^k$, replicated $|A|$ times, for each $A\in \Acal$.
\item $F^k$ is a square diagonal matrix of dimension $\sum_{A \in \Acal} |A|$. Its diagonal consists of $f(A)$, replicated $|A|$ times, for each $A\in \Acal$.
\item $M$ is a matrix of dimension $P\times\sum_{A \in \Acal} |A|$ that ensures $w^k = Mv^k$.
\end{itemize}

\begin{equation*}
\begin{aligned}
\log & \ p(y^k|f)\\
&=   \log \int_{\RR^P} p(y^k| \{v_A^k\}_{A\in \Acal})\prod_{A\in\Acal} p(v_A^k| f(A))dv_A^k\\
&= \log \int_{\RR^P} \Ncal(y^k|X^kMv^k,{\sigma^k}^2I) \prod_{A\in\Acal} \sup_{\zeta_A^k \geq 0}  f(A)^{\frac{|A|}{2}}\Big[  \Ncal\Big(v_A^k| 0,\frac{\zeta_A^k}{f(A)} I\Big) \Big(2\pi\frac{\zeta_A^k}{f(A)}\Big)^{\frac{|A|}{2}} e^{-\phi_A(\zeta_A^k)} \Big]  \prod_{A\in\Acal} dv_A^k\\
&= \log \int_{\RR^P} \sup_{\zeta_A^k \geq 0, A\in \Acal} \Ncal(y^k|X^kMv^k,{\sigma^k}^2I) \prod_{A\in\Acal}  f(A)^{\frac{|A|}{2}} \Ncal\Big(v_A^k| 0,\frac{\zeta_A^k}{f(A)} I\Big) \Big(2\pi\frac{\zeta_A^k}{f(A)}\Big)^{\frac{|A|}{2}} e^{-\phi_A(\zeta_A^k)}  \prod_{A\in\Acal} dv_A^k\\
&\geq \log \sup_{\zeta_A^k \geq 0, A\in \Acal} \int_{\RR^P} \Ncal(y^k|X^kMv^k,{\sigma^k}^2I) \prod_{A\in\Acal}  f(A)^{\frac{|A|}{2}} \Ncal\Big(v_A^k| 0,\frac{\zeta_A^k}{f(A)} I\Big) \Big(2\pi\frac{\zeta_A^k}{f(A)}\Big)^{\frac{|A|}{2}} e^{-\phi_A(\zeta_A^k)}  \prod_{A\in\Acal} dv_A^k\\
&= \sup_{\zeta_A^k \geq 0, A\in \Acal} \log\int_{\RR^P} \Ncal(y^k|X^kMv^k,{\sigma^k}^2I) \prod_{A\in\Acal}  f(A)^{\frac{|A|}{2}} \Ncal\Big(v_A^k| 0,\frac{\zeta_A^k}{f(A)} I\Big) \Big(2\pi\frac{\zeta_A^k}{f(A)}\Big)^{\frac{|A|}{2}} e^{-\phi_A(\zeta_A^k)}  \prod_{A\in\Acal} dv_A^k\\
&=  \sup_{\zeta_A^k \geq 0, A \in \Acal}  \Big\{\log \int_{\RR^P} \Ncal(y^k|X^kMv^k,{\sigma^k}^2I) \prod_{A\in\Acal} \Ncal\Big(v_A^k| 0,\frac{\zeta_A^k}{f(A)} I\Big) \prod_{A\in\Acal} dv_A^k\\
&\qquad \qquad \qquad + \sum_{A\in\Acal}\log \Big[ f(A)^{\frac{|A|}{2}}\Big(2\pi\frac{\zeta_A^k}{f(A)}\Big)^{\frac{|A|}{2}} e^{-\phi_A(\zeta_A^k)}  \Big]\Big\}\\
&=  \sup_{\zeta_A^k \geq 0, A \in \Acal}  \Big\{\log \int_{\RR^P} \Ncal(y^k|X^kMv^k,{\sigma^k}^2I) \Ncal\Big(v^k| 0, Z^kF^{-1} \Big) dv^k \\
&\qquad \qquad \qquad+ \sum_{A\in\Acal} \Big[ \frac{|A|}{2}\log f(A) + \frac{|A|}{2}\log\Big(2\pi\frac{\zeta_A^k}{f(A)}\Big) -\phi_A(\zeta_A^k)  \Big]\Big\}\\
&=  \sup_{\zeta_A^k \geq 0, A \in \Acal}  \Big\{\log \Ncal(y^k|0, X^kM  Z^kF^{-1} M^{\top}{X^k}^{\top} + \sigma^2I) \\
&\qquad \qquad \qquad+ \sum_{A\in\Acal} \Big[ \frac{|A|}{2}\log f(A) + \frac{|A|}{2}\log\Bigl(2\pi\frac{\zeta_A^k}{f(A)}\Bigr) -\phi_A(\zeta_A^k)  \Big]\Big\}\\
&=  \sup_{\zeta_A^k \geq 0, A \in \Acal} \Bigl\{ - \frac{1}{2} {y^k}^{\top}\!\Bigl(X^kM  Z^kF^{-1} M^{\top}{X^k}^{\top} \!\!\!+\! \sigma^2I\Bigr)^{-1}\! y^k - \frac{1}{2} \log\det\Big(  X^kM  Z^kF^{-1} M^{\top}{X^k}^{\top} \!\!\!+\! \sigma^2I\Bigr) \\
&\qquad \qquad \qquad -\frac{N}{2} \log(2\pi) +\sum_{A\in\Acal}\Bigl[ \frac{|A|}{2}\log f(A) + \frac{|A|}{2}\log\Bigl(2\pi\frac{\zeta_A^k}{f(A)}\Bigr) -\phi_A(\zeta_A^k)  \Bigr]\Bigr\}.\\
\end{aligned}
\end{equation*}

Thus, we need to minimize the following overall bound with respect to $f$ and $\zeta_A^k$ for all $A \in \Acal$ and $k\in\{1,\ldots,K\}$:
\begin{equation*}
\tag{\ref{eq:LBgeneral_sum} revisited}
\begin{aligned}
-\sum_{k=1}^K \Big\{ &- \frac{1}{2} {y^k}^{\top}\Big(X^kM  Z^kF^{-1} M^{\top}{X^k}^{\top} + \sigma^2I\Big)^{-1} y^k - \frac{1}{2} \log\det\Big(X^kM  Z^kF^{-1} M^{\top}{X^k}^{\top} + \sigma^2I\Big) \\
 & + \sum_{A\in\Acal}\frac{|A|}{2}\log f(A) + \frac{\sum_{A\in\Acal}|A|-N}{2}\log(2\pi) + \frac{1}{2}\log\det (Z^kF^{-1} ) -\sum_{A\in\Acal}\phi_A(\zeta_A^k)  \Big\}.\\
\end{aligned}
\end{equation*}

In its form given by \eqref{eq:LBgeneral_sum}, the bound is difficult to optimize. However, we can recognize parts of it as minima of convex functions, which will allow us to design an iterative algorithm with analytic updates, finding a local minimum. In particular, it is not difficult to show that
\begin{align*}
 \frac{1}{2} \log &\det \Big(X^kM  Z^kF^{-1} M^{\top}{X^k}^{\top} + \sigma^2I\Big) -\frac{1}{2}\log\det (Z^kF^{-1} )  \tag{matrix determinant lemma}\\
& =\frac{1}{2} \log\det \Big( M^{\top}{X^k}^{\top} X^kM + \sigma^2 F{Z^k}^{-1} \Big) + \frac{N^k\!\!-\!\!\sum_{A\in\Acal}|A|}{2}\log(\sigma^2) \\ 
&= \inf_{\Lambda^k \succcurlyeq 0} \frac{1}{2} \tr\{ (M^{\top}\!{X^k}^{\top}\!X^kM + \sigma^2 F{Z^k}^{-1}) \Lambda^k \} \!-\! \frac{1}{2}\log \det\Lambda^k \!-\! \frac{1}{2}\!\!\sum_{A\in\Acal}\!|A|+ \frac{N^k\!\!-\!\!\sum_{A\in\Acal}|A|}{2}\log(\sigma^2),\\
\end{align*}
which, with a change of variables $\Sigma^k = \sigma^2\Lambda^k$, is written as
\begin{equation*}
\begin{aligned}
\inf_{\Sigma^k \succcurlyeq 0} \frac{1}{2\sigma^2} \tr\{ (M^{\top}{X^k}^{\top}X^kM)\} +\frac{1}{2}\sum_{A\in\Acal} \frac{\tr\Sigma^k f(A)}{\zeta_A^k}  - \frac{1}{2}\log \det\Sigma^k - \frac{1}{2}\sum_{A\in\Acal}|A|+ \frac{N^k}{2}\log(\sigma^2),\\
\end{aligned}
\end{equation*}
and that
\begin{equation*}
\begin{aligned}
\frac{1}{2} {y^k}^{\top}\Big(  X^kM Z^kF^{-1} M^{\top}{X^k}^{\top} + \sigma^2I\Big)^{-1} y^k = \inf_{v^k} \frac{1}{2\sigma^2}||y^k-X^kMv^k ||^2_2 + \frac{{v^k}^{\top} F{Z^k}^{-1} v^k}{2}.\\
\end{aligned}
\end{equation*}

Thus, from \eqref{eq:LBgeneral_sum}, our optimization problem becomes
\begin{equation*}
\tag{\ref{eq:obj} revisited}
\begin{aligned}
\inf_{\zeta^k\geq 0} \inf_{v^k} \inf_{\Sigma^k\succcurlyeq 0} \sum_{k=1}^K\Big\{&\quad \frac{1}{2\sigma^2} ||y^k-X^kMv^k||^2_2 + \frac{1}{2}  \sum_{A\in\Acal} \frac{f(A)}{\zeta_A^k} (||v_A^k||^2_2 + \tr \Sigma_{AA}^k) \\
&- \frac{1}{2} \log\det \Sigma^k + \frac{N^k}{2}\log(\sigma^2)  + \frac{N^k}{2} \log (2\pi)+ \frac{1}{2\sigma^2} \tr{M^{\top}{X^k}^{\top}X^kM\Sigma^k} -\frac{1}{2}  \sum_{A\in\Acal}|A|\\
&+  \sum_{A\in\Acal} \Big[ -\frac{|A|}{2} \log 2\pi + \phi_A(\zeta_A^k) - \frac{1}{2}|A|\log f(A) \Big]\Big\},\\
\end{aligned}
\end{equation*}
and the updates are
\begin{equation}
\label{eq:updates1}
\begin{aligned}
\Sigma^k & = \sigma^2(M^{\top}{X^k}^{\top}X^kM + \sigma^2 F{Z^k}^{-1} )^{-1}\\
v^k   & = (M^{\top}{X^k}^{\top}X^kM + \sigma^2 F{Z^k}^{-1})^{-1}M^{\top}{X^k}^{\top}y^k\\
\zeta_A^k & = \argmin_{z \geq 0} \phi_A(z ) + \frac{1}{2} \frac{f(A)}{z} (||v_A^k||^2_2 + \tr \Sigma_{AA}^k)\\ 
\sigma^2& = \frac{\sum_{k=1}^K\big\{ ||y^k-X^kMv^k||^2_2 +   \tr M^{\top}{X^k}^{\top}X^kM\Sigma^k \big\}} {\sum_{k=1}^{K}N^k}\\
 f(A) & = \argmin_{x} \frac{x}{2}\sum_{k=1}^K \frac{1}{\zeta_A^k} (||v_A^k||^2_2 + \tr \Sigma_{AA}^k)  - K\frac{1}{2}|A|\log x\\
& = \frac{K|A|}{\sum_{k=1}^K \frac{1}{\zeta_A^k} (||v_A^k||^2_2 + \tr \Sigma_{AA}^k)}. \\
\end{aligned}
\end{equation}

\subsection{Regularized version}
 
As we empirically show in Section 3.1, the variational approximation scheme from above tends to overestimate the variance of the prior distribution (i.e., underestimate the inverse scale parameter $f(A)$) when this variance is smaller than $\sigma^2$, the noise variance. This is undesirable, as we would like $f(A)$ to go to infinity for irrelevant subsets of variables. To circumvent this problem, we use an improper prior of the form 
$$p(f(A)) \propto f(A)^\beta$$ 
to encourage $f(A)$ to go to infinity when the variance of $p(v_A)$ is smaller than $\sigma^2$.
Consequently, the term $-\beta\log f(A)$ is added to the objetive function \eqref{eq:obj}, and the only update that changes is the update for $f(A)$:
\begin{equation}
\begin{aligned}
f(A) & = \argmin_{x} \frac{x}{2}\sum_{k=1}^K \frac{1}{\zeta_A^k} (||v_A^k||^2_2 + \tr \Sigma_{AA}^k)  - K(\frac{|A|}{2}+\beta)\log x\\
& = \frac{K(\frac{|A|}{2}+\beta )}{\frac{1}{2}\sum_{k=1}^K \frac{1}{\zeta_A^k} (||v_A^k||^2_2 + \tr \Sigma_{AA}^k)}.\\
\end{aligned}
\end{equation}

\subsection{Faster updates}
The update equations \eqref{eq:updates1} involve the inversion of the $\sum_{A\in\Acal}|A|\times \sum_{A\in\Acal}|A|$ matrix $M^{\top}{X^k}^{\top}X^kM + \sigma^2 F{Z^k}^{-1}$. 
In fact, using the matrix inversion lemma, we can avoid performing this expensive inversion by rewriting the updates so that we only have to invert a $P\times P$ or an $N^k\times N^k$ matrix instead.

Before we write down the modified updates, let us introduce some additional shorthand notation:
\begin{itemize}
\item $\xi^k \in \RR^P$ is the sum $\sum_{A\in\Acal} \frac{\zeta_A^k}{f(A)} 1_A$, where $1_A\in \RR^P$ denotes the indicator vector for the index set $A$;
\item $\Xi^k \in \RR^{P\times P}$ is a square diagonal matrix with $\Xi^k_{ii}=\xi^k_i, i=1,\ldots,P$; Put differently, $\Xi^k=MZ^kF^{-1}M^{\top}$.
\item $H^k$ is a square diagonal matrix corresponding to $Z^kF^{-1}$.
\end{itemize}

\paragraph{$P\times P$ matrix inversion.}
\begin{equation}
\label{eq:updates_PP}
\begin{aligned}
\Sigma^k & = H^k - H^k M^{\top}  {\Xi^k}^{-1}M H^k +\sigma^2  H^k M^{\top} {\Xi^k}^{-1} \big({X^k}^{\top}  X^k + \sigma^2 {\Xi^k}^{-1}\big)^{-1}{\Xi^k}^{-1} M H^k     \\
\tr\Sigma_{AA}^k & = |A|\frac{\zeta_A^k}{f(A)} - \frac{{\zeta_A^k}^2}{{f(A)}^2}\sum_{i\in A} \frac{1}{\xi^k_i} + \sigma^2 \frac{{\zeta_A^k}^2}{{f(A)}^2} \sum_{i,j\in A} \frac{1}{\xi^k_i \xi^k_j} \big[\big({X^k}^{\top}  X^k + \sigma^2 {\Xi^k}^{-1}\big)^{-1}\big]_{ij} \\
v^k   & =  H^k M^{\top} {\Xi^k}^{-1} \big({X^k}^{\top}  X^k + \sigma^2 {\Xi^k}i^{-1}\big)^{-1}{X^k}^{\top}y^k\\
\|v_A^k\|_2^2   & =  \frac{{\zeta_A^k}^2}{{f(A)}^2} \sum_{i \in A} \frac{1}{{\xi^k_i}^2} \big[\big({X^k}^{\top}  X^k + \sigma^2 {\Xi^k}^{-1}\big)^{-1} {X^k}^{\top}y^k\big]_i^2.\\
\end{aligned}
\end{equation}

\paragraph{$N^k\times N^k$ matrix inversion.}
\begin{equation}
\label{eq:updates_NN}
\begin{aligned}
\Sigma^k & = H^k - H^k M^{\top}  {X^k}^{\top} \big( X^k \Xi^k {X^k}^{\top} + \sigma^2 I\big)^{-1} X^kM H^k     \\
v^k   & = H^k M^{\top}{X^k}^{\top} \big( X^k \Xi^k {X^k}^{\top} + \sigma^2 I\big)^{-1} y^k.\\
\end{aligned}
\end{equation}

\paragraph{Special case of no design (signal denoising).}
When $X^k=I$, the computations become considerably simpler. Note that in this case $N^k=P$ and the matrix ${X^k}^{\top}  X^k + \sigma^2 {\Xi^k}^{-1}$ is diagonal, so the cost of its inversion is $O(P)$ instead of $O(P^3)$. In fact, we do not even need to form the diagonal matrix, as we do not need to explicitly use $\Sigma^k$ and $v^k$ in the updates. The updates can be rewritten as follows: 
\begin{equation}
\label{eq:updates_nodesign}
\begin{aligned}
\tr\Sigma_{AA}^k & = |A|\frac{\zeta_A^k}{f(A)} - \frac{{\zeta_A^k}^2}{{f(A)}^2}\sum_{i\in A} \frac{1}{\xi^k_i} + \sigma^2 \frac{{\zeta_A^k}^2}{{f(A)}^2} \sum_{i\in A} \frac{1}{{\xi^k_i}^2}\frac{1}{(1+\sigma^2/\xi^k_i)} \\
\|v_A^k\|_2^2   & =  \frac{{\zeta_A^k}^2}{{f(A)}^2} \sum_{i \in A} \frac{1}{{\xi^k_i}^2} \Big[\frac{y_i^k}{1+\sigma^2/\xi^k_i}\Big]^2\\
\zeta_A^k & = \argmin_{z \geq 0} \phi_A(z ) + \frac{1}{2} \frac{f(A)}{z} (||v_A^k||^2_2 + \tr \Sigma_{AA}^k)\\ 
 f(A) & = \frac{K(\frac{|A|}{2}+\beta )}{\frac{1}{2}\sum_{k=1}^K \frac{1}{\zeta_A^k} (||v_A^k||^2_2 + \tr \Sigma_{AA}^k)}.\\
\end{aligned}
\end{equation}
(The updates for $f(A)$ and $\zeta_A^k$ remain unchanged.)

In this case the computation of $w^k, k\in\{1,...,K\}$, and of the value of the objective are also simplified. To obtain $w^k= \sum_{A \in \Acal} v_A^k$, we compute each component of $v_A^k$ as 
\begin{equation*}
\begin{aligned}\quad
[v_A^k]_i  =  \frac{{\zeta_A^k}}{{f(A)}} \frac{1}{\xi^k_i} \frac{y_i^k}{1+\sigma^2/\xi^k_i}  =\frac{{\zeta_A^k}}{{f(A)}} \frac{y_i^k}{\xi^k_i+\sigma^2} \mbox{ if } i\in A, 0 \mbox{ otherwise,}\\
\end{aligned}
\end{equation*}
and the objective as
\begin{equation*}
\begin{aligned}
\inf_{\zeta^k\geq 0} \inf_{v^k} \inf_{\Sigma^k\succcurlyeq 0} \sum_{k=1}^K\Big\{&\quad \frac{1}{2\sigma^2} ||y^k-w^k||^2_2 + \frac{1}{2}  \sum_{A\in\Acal} \frac{f(A)}{\zeta_A^k} (||v_A^k||^2_2 + \tr \Sigma_{AA}^k) \\
&+ \frac{1}{2} \sum_{i=1}^P \log(\sigma^2 + \xi^k_i) +\frac{1}{2} \sum_{A \in \Acal} |A| \log\frac{f(A)}{\zeta_A^k}  \\
&+ \frac{P}{2} \log (2\pi) + \frac{1}{2}  \sum_{i=1}^P \frac{\xi^k_i}{\xi^k_i + \sigma^2}  -\frac{1}{2}  \sum_{A\in\Acal}|A|\\
& +  \sum_{A\in\Acal} \Big[ -\frac{|A|}{2} \log 2\pi + \phi_A(\zeta_A^k) - \frac{1}{2}|A|\log f(A) \Big]\Big\}.\\
\end{aligned}
\end{equation*}
Here we have used 
\begin{equation*}
\begin{aligned}
 \log\det \Sigma^k & =   \log\det [\sigma^2 (M^{\top} M + \sigma^2F{Z^k}^{-1})^{-1}] \\
& =  \sum_{A\in \Acal}|A|\log\sigma^2 - \log\det (M^{\top} M + \sigma^2 F{Z^k}^{-1}) \\
& = \sum_{A\in \Acal}|A|\log\sigma^2 -  \log[  {\sigma^2}^{\sum_{A\in\Acal}|A|-P} \det (\sigma^2I + M^{\top} Z^kF^{-1}M) \det (F{Z^k}^{-1})]\\
& =P\log\sigma^2  -\sum_{i=1}^P \log(\sigma^2 + \xi^k_i)-\sum_{A \in \Acal} |A| \log\frac{f(A)}{\zeta_A^k}\\
\end{aligned}
\end{equation*}
and
\begin{equation*}
\begin{aligned}
\tr {M^{\top}M\Sigma^k} &=  \tr {M\Sigma^kM^{\top}}\\
& =\tr{MH^kM^{\top}} -\tr{ MH^kM^{\top}  (\Xi^k + \sigma^2I)^{-1} MH^kM^{\top} }\\
& =\tr{\Xi^k} -\tr{ \Xi^k  (\Xi^k + \sigma^2I)^{-1} \Xi^k }\\
& =\sum_{i=1}^P\xi^k_i - \sum_{i=1}^P \frac{{\xi^k_i}^2}{\xi^k_i + \sigma^2}\\
&=\sum_{i=1}^P \frac{\xi^k_i \sigma^2}{\xi^k_i + \sigma^2}.\\
\end{aligned}
\end{equation*}
Note that if we update the variables in the same order as in \eqref{eq:updates_nodesign}, then $w^k$, $\log\det \Sigma^k$, and $\tr {M^{\top}M\Sigma^k}$ have to be computed before updating $\zeta_A^k$ and $f(A)$; This will ensure that $w^k$ and $||v_A^k||^2_2$, respectively $\tr \Sigma_{AA}^k$ and the two terms involving $\log\det \Sigma^k$ and $\tr {M^{\top}M\Sigma^k}$, are consistent, that is, they correspond to the same value of $v_A^k, A\in\Acal$, respectively $\Sigma^k$.
}

\section*{Acknowledgements}
This work was supported by the European Research Council (SIERRA project 239993 and SMAC project 280032). The authors would like to thank Guillaume Obozinski for fruitful discussions, Julien Mairal for his advice on setting up experiments on images, and Sylvain Arlot for his comments on the manuscript. A large part of this work was done while Nino Shervashidze was also affiliated with the Sierra project-team at INRIA and \'Ecole Normale Sup\'erieure.

\bibliography{refs}
\bibliographystyle{abbrvnamed}
\end{document}